\documentclass{llncs} 
\usepackage{times}
\usepackage{amsmath} 
\usepackage{helvet}
\usepackage{courier}
\usepackage{graphicx}
\usepackage{algorithm}
\usepackage[noend]{algpseudocode}
\usepackage{needspace}

\begin{document}
\title{Deep Interactive Evolution}
\author{Philip Bontrager$^{1}$, Wending Lin$^{3}$, Julian Togelius$^{1}$ and Sebastian Risi$^{2}$}

\institute{New York University 
	\\New York, United States\\  
    \email{philibjb@nyu.edu, julian@togelius.com} 
\and {IT University of Copenhagen 
	\\Copenhagen, Denmark\\
	\email{sebr@itu.dk}} 
\and {Beijing University of Posts and Telecommunications 
	\\Beijing, China\\
	\email{wendinglin96@gmail.com}}}
\authorrunning{P. Bontrager et al.}

\maketitle

\begin{abstract}
This paper describes an approach that combines \emph{generative adversarial networks} (GANs) with interactive evolutionary computation (IEC). While GANs can be trained to produce lifelike images, they are normally sampled randomly from the learned distribution, providing limited control over the resulting output. On the other hand, interactive evolution has shown promise in creating various artifacts such as images, music and 3D objects, but traditionally relies on a hand-designed evolvable representation of the target domain. The main insight in this paper is that a GAN trained on a specific target domain can act as a compact and \emph{robust} genotype-to-phenotype mapping (i.e.\ most produced phenotypes do resemble valid domain artifacts). Once such a GAN is trained, the latent vector given as input to the GAN's generator network can be put under evolutionary control, allowing controllable and high-quality image generation.   In this paper, we demonstrate the advantage of this novel approach through a user study in which participants were able to evolve images that strongly resemble specific target images.
\end{abstract}

\section{Introduction}

This paper addresses the question how to generate artifacts, included but not limited to images, through indirect interaction with computer software. This means that instead of effecting change directly to the artifact, for example through drawing a picture with virtual brushes in a traditional image editing software, the user communicates with the software through for example evaluative feedback and suggestions.

Such interactive or AI-assisted creation has a number of potential uses. Perhaps most obviously, combining the generative capacity of computer software with human input can allow humans to create artifacts (such as images) of a quality and richness they could not create on their own using direct methods because they do not possess the requisite technical skills (such as drawing). But such methods could also be used to explore or enforce particular aesthetics, allowing users to create art in some other aesthetic than they would normally adhere to. Use cases for such AI-assisted creation methods could include creating avatars for games and virtual environments, creating facial composites for help in criminal cases, and product design, customization or search.

Interactive evolution is a particular form of AI-assisted creation where the human user functions as the fitness function in an evolutionary algorithm. The system serves up a number of artifacts to the human every generation, and the human responds by indicating which of these artifact(s) they prefer. Through crossover and/or mutation, the system then generates another generation of artifacts, from which the user selects their favorite artifact(s), and so on~\cite{takagi2001interactive}.

While the conceptual appeal of interactive evolution is strong, there are several problems hampering the practical usefulness of the technique. One issue is that a large number of evaluations may be necessary to find the desired artifact, leading to an issue known as \emph{user fatigue}. It might also not be possible to find the desired artifact through optimizing for traits that seem to be correlated with, or found in, the desired artifact; some studies suggest that such ``goal-directed evolution'' often fails~\cite{secretan2008picbreeder,woolley2011deleterious}. Part of the problem here might be the underlying representation and genotype-to-phenotype mapping employed in the interactive evolution system, which might not be conducive to search, or which might thread the wrong balance between generality and domain-specificity.

One potential solution is to find better artifact representations, which can map between genotype and phenotype within a particular domain while preserving the property that most points in search space correspond to reasonable artifacts. We suggest that this could be done by using the generator parts of Generative Adversarial Networks (GANs), which are trained to produce images (or other artifacts) from a particular domain. We hypothesize that using a generative representation acquired through adversarial training can give us a searchable domain-specific space. We call this approach \emph{deep interactive evolution}.

In the rest of this paper, we discuss previous work in interactive evolution and generative adversarial networks and present our specific approach. We also show the results of letting a set of users use deep interactive evolution using a generator trained on a shoe dataset, and one trained on a face dataset.

\section{Background}

This section reviews interactive evolutionary and generative adversarial networks, which are fundamental for the combined approach presented in this paper. 

\subsection{Interactive Evolutionary Computation (IEC)}

In \emph{interactive evolutionary computation} (IEC) the traditional objective fitness function is replaced by a human selecting the candidates for the next generation. 
IEC has traditionally been used in optimization tasks of subjective criteria or in open-ended domains where the objective is undefined \cite{Pallez2007,takagi2001interactive,todd1992evolutionary}.  
As surveyed by Takagi~\cite{takagi2001interactive}, IEC has been applied to several domains including image generation \cite{sims1991artificial}, games \cite{gar2009,petalz2012}, music \cite{hoover2011interactively}, industrial design, and data mining. Evolutionary art and music often hinges on human evaluation of content due to the subjective nature of aesthetics and beauty.

Most IEC approaches follow the paradigm of the original Blind Watchmaker from Dawkins \cite{dawkins1986blind}, in which artifacts are evolved by users through an interactive interface that allows them to iteratively select from a set of candidate designs. However, a problem with this IEC approach is \textit{user fatigue}, and many attempts have been made to minimize it \cite{Bongard2013,Kamalian2005,Pallez2007}. User fatigue is a problem because evolution takes time and humans tend to suffer from fatigue after evaluating relatively few generations. 

One approach to limit user fatigue is \emph{collaborative interactive evolution}, in which users can continue evolution from promising starting point generated by other users. Picbreeder\footnote{http://picbreeder.org/} is an example of such a system, which allows users to collaboratively evolve images online, building on the intermediate results published by others \cite{secretan2008picbreeder}. Another approach is to seed the initial population
of IEC with meaningful, high-quality individuals \cite{zhang2015drawcompileevolve}.

In this paper we introduce a novel approach that should suffer less from user fatigue by restricting the space of possible artifacts to a certain class of images (e.g.\ faces, shoes, etc.); the hypothesis is that this way the user needs to spend less time
exploring the space of possible solutions. 

\subsection{Generative Adversarial Networks (GANs)}
\begin{algorithm}
\caption{Generative Adversarial Network}
\label{alg:GAN}
\begin{algorithmic}[1]
\State Initialize $\theta$ and $\omega$
\For {t training iterations}
	\For {k discriminator updates}
		\State $x \gets $ Minibatch sample of real data
    	\State $z \gets $ Minibatch of latent variables where $z \sim p(z)$
        \State $L \gets Loss(D_\omega(x), Real) + Loss(D_\omega(G_\theta(z)), Generated)$
    	\State $\omega \gets $ gradient-based update w.r.t L
    \EndFor
    \State $z \gets $ Minibatch of latent variables where $z \sim p(z)$
    \State $L \gets Loss(D_\omega(G_\theta(z)), Real)$
    \State $\theta \gets $ gradient-based update w.r.t L
\EndFor
\end{algorithmic}
\end{algorithm}

The space of potential images in our approach is represented by a Generative and Adversarial Network (GAN). GANs \cite{goodfellow2014generative} are a new class of deep learning algorithms that have extended the reach of deep learning methods to unsupervised domains, which do not require
labeled training data. The basic idea behind GANs is that two networks compete against each other: one network is generative and the other network is discriminative. The generative network tries to create synthetic content (e.g.\ images of faces) that is indistinguishable to the discriminative network from real data (e.g.\ images of real faces). Both the generator and discriminator are trained in turns, with the generator becoming better at forging content, and the discriminator becoming better at telling real from synthetic content.

In more detail, a generator network (called G) conditioned on latent variables, generates images to convince a separate discriminator network (called D) that its generated images are authentic ones from a real-world training set. Algorithm \ref{alg:GAN} outlines the basic steps of training a generator using this adversarial technique. Typically the discriminator is only updated once for every time the generator is updated, so k is 1. A lot of research on GANs is now focused on what loss functions allow for the most stable training. The latent variables are random variables that are typically sampled from a normal distribution. A generator is trained on a fixed number of latent variables and the stochasticity forces the generator to generalize.

\section{Approach: Deep Interactive Evolution}
\begin{figure*}
\begin{center}
  \includegraphics[width=\textwidth]{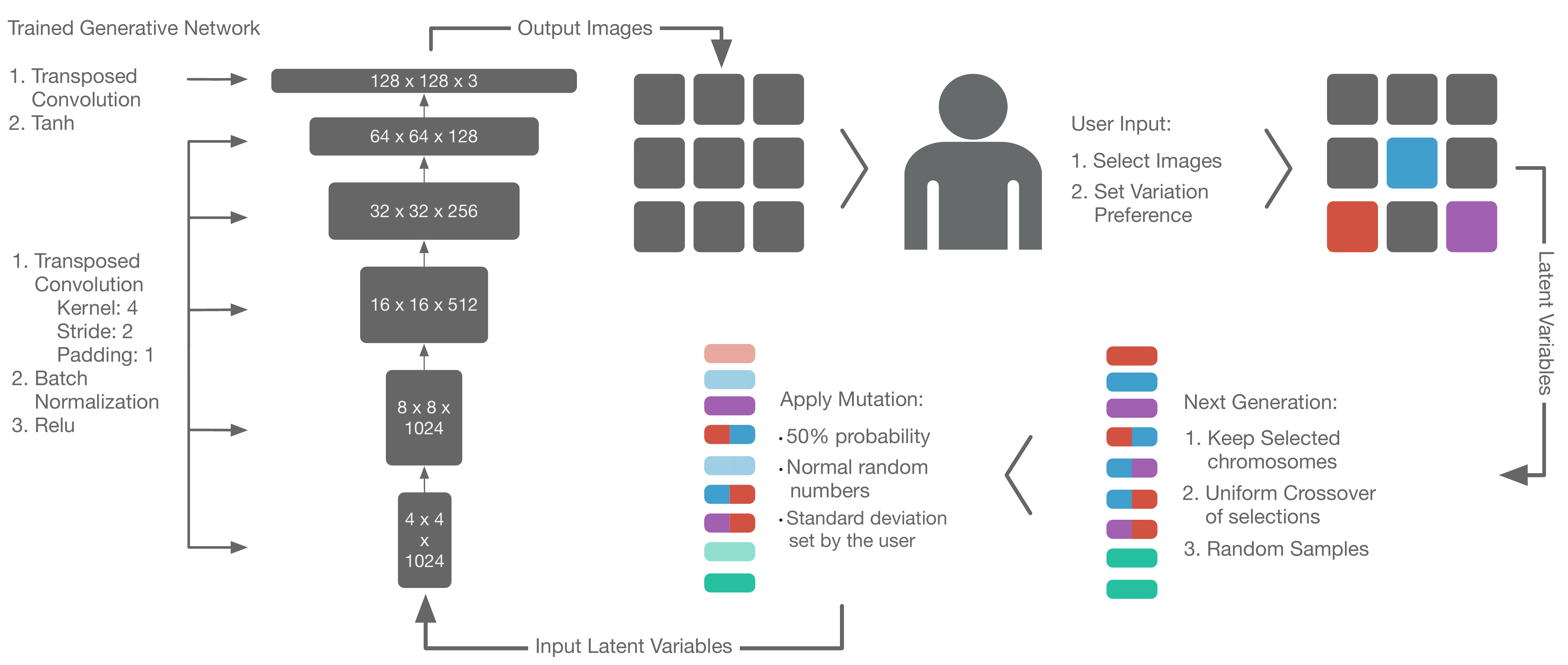} 
  \caption{This figure illustrates the four steps of DeepIE. First, latent variables are passed through a trained image generator to produce images. Next, the user selects the images based on their interests. Third, new sets of latent variables are chosen based on the selected image. Last, the latent variables are mutated with the intensity set by the user.} 
  \label{fig:algo}
  \vspace{-1.0em}
\end{center}
\end{figure*}
Figure~\ref{fig:algo} depicts our Deep Interactive Evolution (DeepIE) approach. The main differentiating factor between DeepIE and other interactive evolution techniques is the employed generator. The content generator is trained over a dataset to constrain and enhance what is being evolved. In the implementation in this paper,  we trained a non-specialized network over 2D images. In general, a number of goals can be optimized for during the training process. For example, for generating art, a network that specializes in creative output can be used. Recently Elgamma et. al. proposed CAN, a variation to the GAN training that prioritizes creative output \cite{elgammal2017can}. In other work, Kim et. al. train a network to map latent variables to discrete outputs, such as text \cite{kim2017adversarially}.

The generator works by transforming a number of latent variables into content. In our work, the generator has 20 latent variables as input and the content is images from various domains. Typically the latent variables are chosen at random to produce random, pleasing, images. In DeepIE, we initially sample the latent variables randomly and then evolve them to get the generator to produce the user's optimal image.

The randomly sampled latent variables are used to generate images. The best-generated images are then selected by the user. Then the latent variables that produced the images are evolved to create a set of new variables. 

There are many ways this can be approached. The user could simply have the ability to select images or the complexity could be increased and the user could be allowed to provide more information such as the ability to rank the images or score them. Depending on the interface, an evolutionary strategy can be chosen to take advantage of the provided information. It is important to use crossover as it allows the user the ability to guide the process through picking images that can be later combined. The user then determines the amount of mutation to apply which can be implemented in a number of ways.

This setup is very general. To create images in a specific domain, the person just needs a sufficient amount of data to train the generator. Once the generator is automatically trained on a domain, the user can evolve the inputs exactly the way they have always done. To use DeepIE on content outside of images, the generator architecture might need to be modified but the algorithm still remains the same.

\subsection{Experimental Setup}

\begin{algorithm}[tbh]
\caption{Deep Interactive Evolution.  \newline
Defaults $m, n \gets 20, \mu \gets 0, \sigma^2 \gets 1, p \gets .5, foreign \gets 2$}
\label{alg:deepIE}
\begin{algorithmic}[1]
\State $G_\theta \gets \textit{trainGAN}(data)$%
\Comment{\parbox[t]{.35\linewidth}{we use WGAN-GP \cite{gulrajani2017improved}}}
\State $interface \gets \text{Interface()}$
\State $Z \gets \text{m by n matrix where } z_{i,j} \sim \mathcal{N}(\mu, \sigma^2)$\Comment{\parbox[t]{.35\linewidth}{population size m with n latent variables each}}
\Repeat{}
	\State	$images \gets G_\theta(Z)$
	\State $\text{interface.display}(images)$
	\State $\textbf{wait until } \text{interface.buttonNextPressed() is True}$
	\State $indices \gets \text{interface.getSelectedImages()}$
	\State $selection \gets Z_{indicies}$
	\State $s^2 \gets \text{interface.getMutationParameter()}$
	\State $\Delta \gets m - length(selection)$
	\State $x \gets \max(0, \Delta - foreign)$
	\State $cross \gets x \text{ by n matrix where } cross_{i} \sim mutate(uniform(selection), s^2)$
	\State $x \gets \max(foreign, \Delta)$
	\State $new \gets x \text{ by n matrix where } new_{i, j} \sim \mathcal{N}(\mu, \sigma^2)$
	\State $selection \gets \text{apply } mutate \text{ with } s^2 \text{ to each } selection_i$
	\State $Z \gets selection + cross + new$
\Until{user stops}

\Function{uniform}{population}
	\State $a \gets$ random individual from population
	\State $b \gets$ random individual from population
	\State $mask \gets \text{vector of length n where } mask_i \sim Bernoulli(.5)$
	
	\Return $mask\cdot a + (1-mask)\cdot b$
\EndFunction

\Function{mutate}{individual, std}
	\State $mutation \gets Bernoulli(p)$
	\State $noise \gets \text{vector of length n where} noise_i \sim \mathcal{N}(\mu, std)$ 
	
	\Return $individual + mutation\cdot noise$
\EndFunction
\end{algorithmic}
\end{algorithm}

In this work, we implement a vanilla version of DeepIE. This allows us to observe the effectiveness of the technique in the most general sense. Here we use a fairly simple setup for the evolutionary process. Also, the interface is kept minimalistic to try to keep the focus off of how different interface design could affect the usability of the system. More involved interfaces could allow more information to be shared with the evolutionary algorithm allowing us to tune the process for different people and domains. This can provide more information but also may increase user fatigue. Testing this is beyond the scope of this paper.

\subsubsection{Generator}

The recent advances in neural network image generators are what enable this new take on interactive evolutionary computation to work. There are a number of techniques for designing and training a data generator. These include GANs, variational autoencoders, and autoregressive techniques, as well as a few more techniques \cite{goodfellow2016nips}. For our setup, we decided to implement a GAN. GANs have seen a lot of recent success in generating realistic looking, relatively high resolution, images.

At the time of publishing this paper, one of the state of the art techniques for training a GAN on images is the Wasserstein GAN with Gradient Penalty (WGAN-GP) \cite{gulrajani2017improved}. This process uses a Wasserstein distance function as a loss for the discriminator and gradient penalties to keep the gradients in an acceptable range for the loss function. The generator has been shown to produce even better results if the discriminator is made to classify the data as well. In this work, we trained our network with the default setup for WGAN-GP without classification to keep our setup as general as possible.

This training method is typically combined with a deep convolutional network architecture (DCGAN) \cite{radford2015unsupervised}. Our implementation of the network is shown in Figure~\ref{fig:algo} on the left end. The generator network is made of repeating modules that consist of a transposed convolution, a batch normalization layer, and a ReLU nonlinearity function. The discriminator is the same architecture upside-down, each module consists of a convolutional layer and a LeakyReLU nonlinearity. The convolutional layer has a kernel of size 3 and stride of 2 for downsampling. To get good consistency from the batch normalization layers during training and evaluation, it’s better to train with large batch sizes.

To minimize the number of variables to evolve, our network only has 20 latent variables. This means the input to G is a random normal vector of size 20. Berthelot et. al. did tests in their work on Boundary Equilibrium GANs and found that changing the number of latent variables did not have a noticeable impact on the diversity of the network \cite{berthelot2017began}. 

\needspace{5\baselineskip}
\subsubsection{Evolution}

The implemented evolutionary process consists of mutation and crossover as the two primary operators for optimization. Mutation allows for local variation of the user's selections. Crossover is what really allows the user to guide the process. The user can select images that do not match what they are looking for but contain a feature they want in their final design. Crossover allows the possibility of that feature being added to the current best design.

To keep the number of generations low, it is important to have frequent mutations to quickly sample a design space. We have 50\% probability of mutation. For mutation, we generate a random normal vector of the same size as the latent variable vector and add it to the latent variables. The standard deviation is set by the user. At the low end, it is set to 0 which results in no mutation. At the high end, it is 1 which is the same magnitude as the latent variables themselves.

\sloppy We use uniform crossover to combine selected parents. A new chromosome is created by randomly sampling from the variables of two parents, $child_i = \allowbreak RandomChoice(parent^A_i, parent^B_i)$ for each variable $i$. Since the variables are not independent in their effect on the generator, this form of crossover can lead to some unexpected results. Testing does show some independence though as features of two images will combine in practice.

Since the user only selects the images they want to keep, the system has no way of evaluating the relative quality or ranking of the selected images. The selected images are kept. The population is rebuilt through random crossover and two new foreign chromosomes. The new chromosomes are added to continually provide the user with new features to select. Our population size is 20 to keep from being overwhelming.

We are aware that using these mutation and crossover techniques result in latent variables that do not match the prior distribution of variables that the network, G, was trained on. This is likely the cause of increased image artifacts the longer the images are evolved. Other options for mutation and crossover could involve interpolating between vectors along the hypersphere to maintain the expected distribution of inputs for G.  

\subsubsection{Data}

We decided to produce 2D images as there is a lot of recent advances in using networks to generate images. We looked for datasets that represent domains that are interesting to work in. In our initial setup, we use three datasets: CelebA face dataset \cite{liu2015faceattributes}, UT Zappos50K shoes \cite{finegrained}, and 3D Chairs images \cite{Aubry14}. A separate network is trained for each domain. To add a new domain, one just has to apply a new dataset. In theory, this setup should translate to 3D data with very few alterations.

\subsubsection{Interface}

\begin{figure*}
\begin{center}
	  \includegraphics[width=\textwidth]{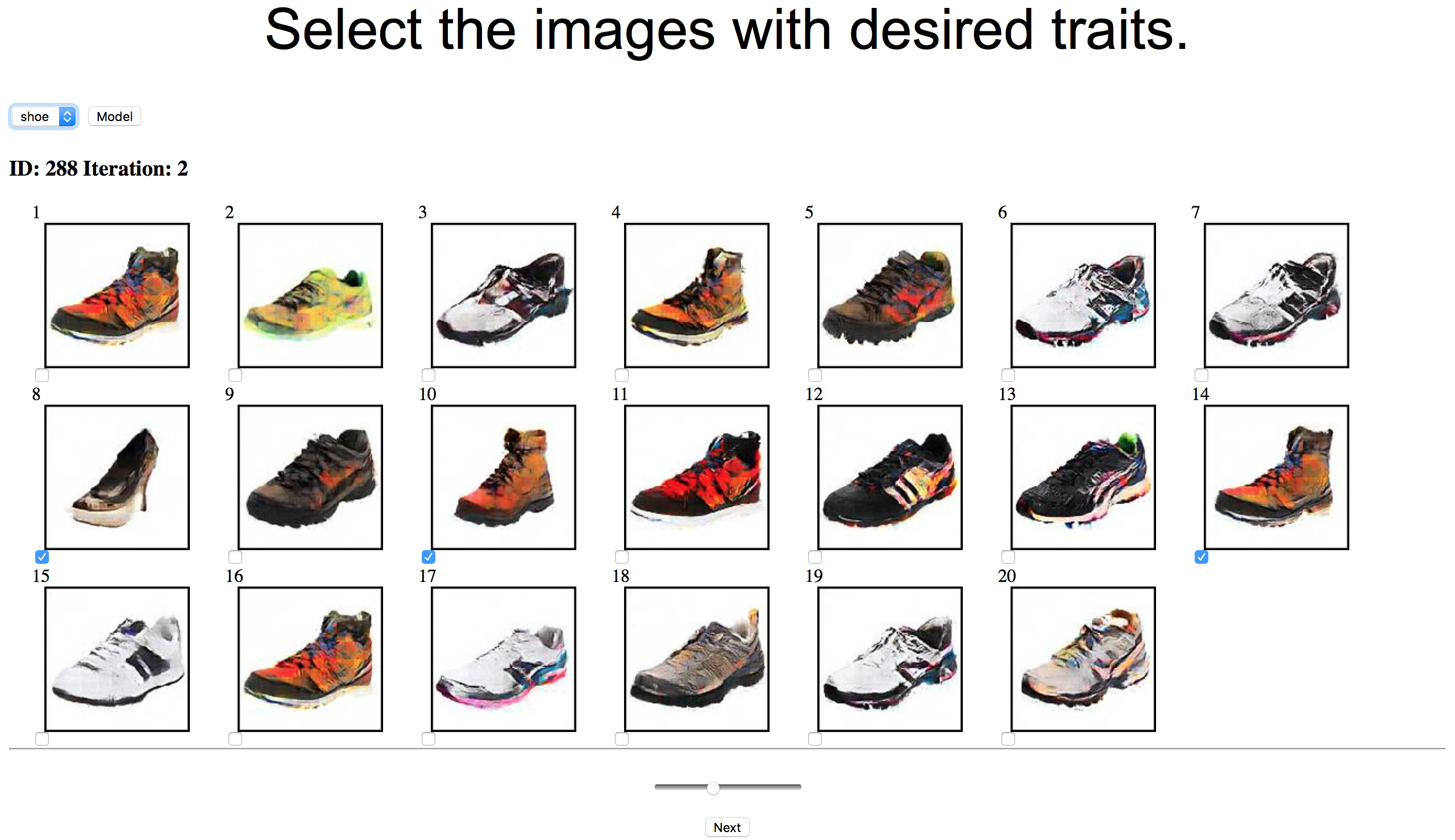} 
  \caption{Web interface for our DeepIE framework.} 
  \label{fig:interface}
  \vspace{-1.0em}
\end{center}
\end{figure*}

Figure~\ref{fig:interface} shows a screenshot of our implementation of DeepIE. There is a dropdown menu at the top left that allows the user to select the domain to design in. They are then presented with the entire generation. The user simply selects which images will go to the next generation. At the bottom is a slider that allows the user to set the standard deviation of the noise that is added during mutation. The user only has to know that further right on the slider increases variability and further left decreases it.

\subsection{User Testing}

To get feedback on our implementation of DeepIE, we had volunteers try the interface. We gave them two main tasks:
\begin{enumerate}
  \item Reproduce the image of a shoe the was created in our system.
  \item Reproduce the image of a face of their choosing.
\end{enumerate}
The intention is to see how much control the user has over the evolution process. While a major use case for this tool is to help someone communicate an image that only exists in their head, for the purpose of testing we required a target image to be chosen in advance.

The volunteers were asked to familiarize themselves with the system by figuring out how to evolve a chair. Once they were comfortable, they were asked to select a shoe image from a list of generated images. They then were asked to use a minimum of 10 generations to try and recreate that image.

The volunteers were given more freedom during the second task. They could choose any public domain face image, or their own, to recreate. Once again they were asked to use a minimum of 10 generations.

After they completed these tasks we asked them to rate their ability to do each of the two tasks on a scale from 1 through 5. We also asked them to describe their strategies for creating the image as well as their experience.

After they finished answering these questions we showed them a randomized mix of all the shoes they selected, and of all the faces they selected and asked them to pick the best from each domain. In this way, we could get a crude measure of whether the images were improving over time.

\section{Results}

\begin{figure*}
\begin{center}
  \includegraphics[width=\textwidth]{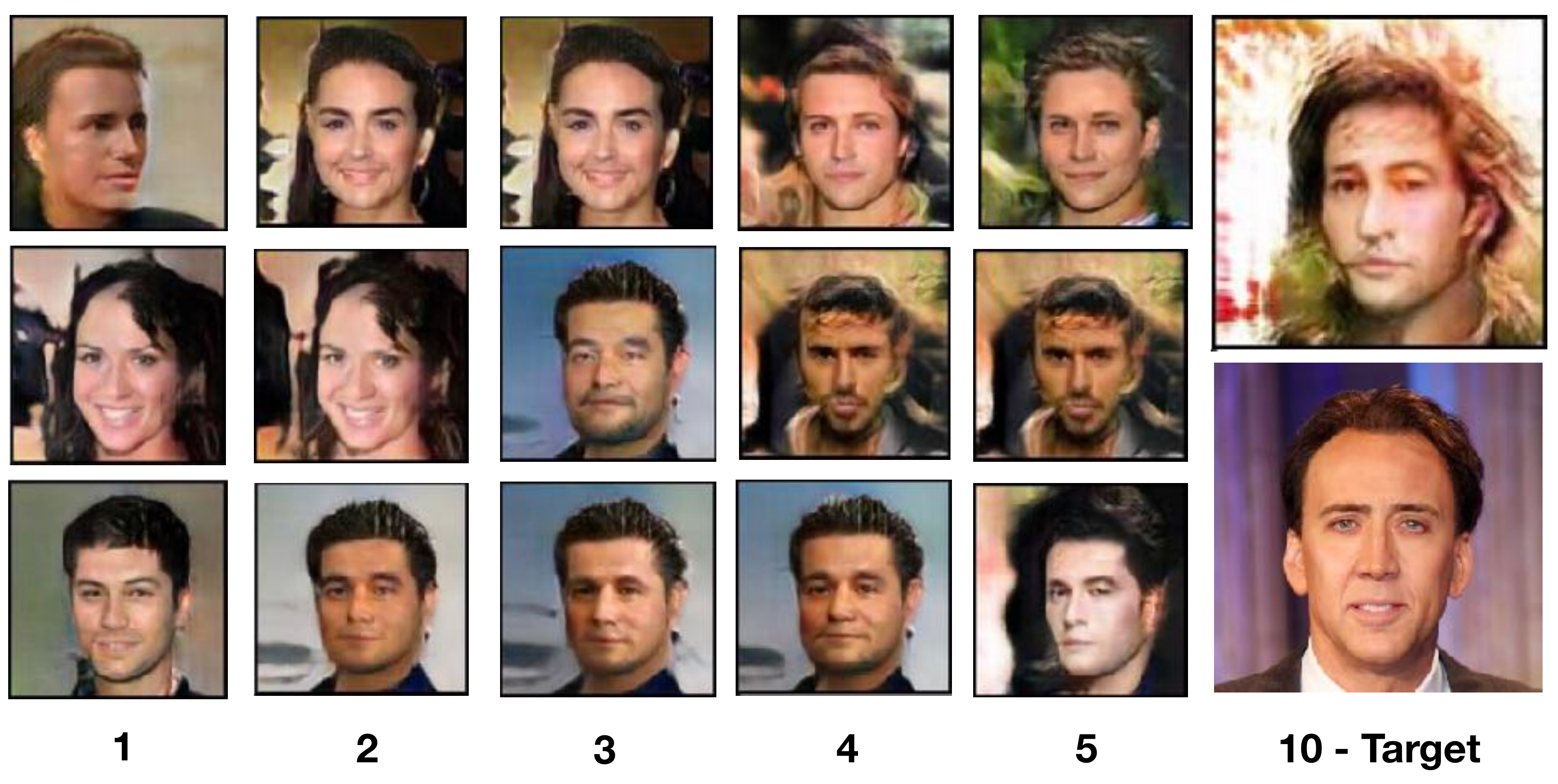} 
  \caption{This demonstrates the authors' attempt to evolve an image that looks like Nicolas Cage.  Each column contains the selected images for the given generation. In this example three images were selected at each step (it is possible to select less or more images). The last column shows the best image after 10 generations.} 
  \label{fig:cage}
  \vspace{-1.0em}
\end{center}
\end{figure*}

The output of our trained network is visible in the images we provided. Due to difficulties in hosting a PyTorch enabled website, we are not currently publicly hosting our site. Figure~\ref{fig:cage} shows the process of evolving an image of Nicholas Cage through selection. Each column represents the selected images for the marked generation. In the example given, each selection is made based on a particular feature it possesses; i.e.\ face shape, hair, pose, etc. Over time, two separate tracks that are being evolved will merge in an acceptable way and they can be merged and the original tracks can be let go. This can be seen between generation 3 and 4 in Figure~\ref{fig:cage} where the top and middle image from generation 3 created the middle image in generation 4.

\subsection{User Testing}

\begin{figure}
\begin{center}
  \includegraphics[width=\textwidth,height=0.88\textheight,keepaspectratio]{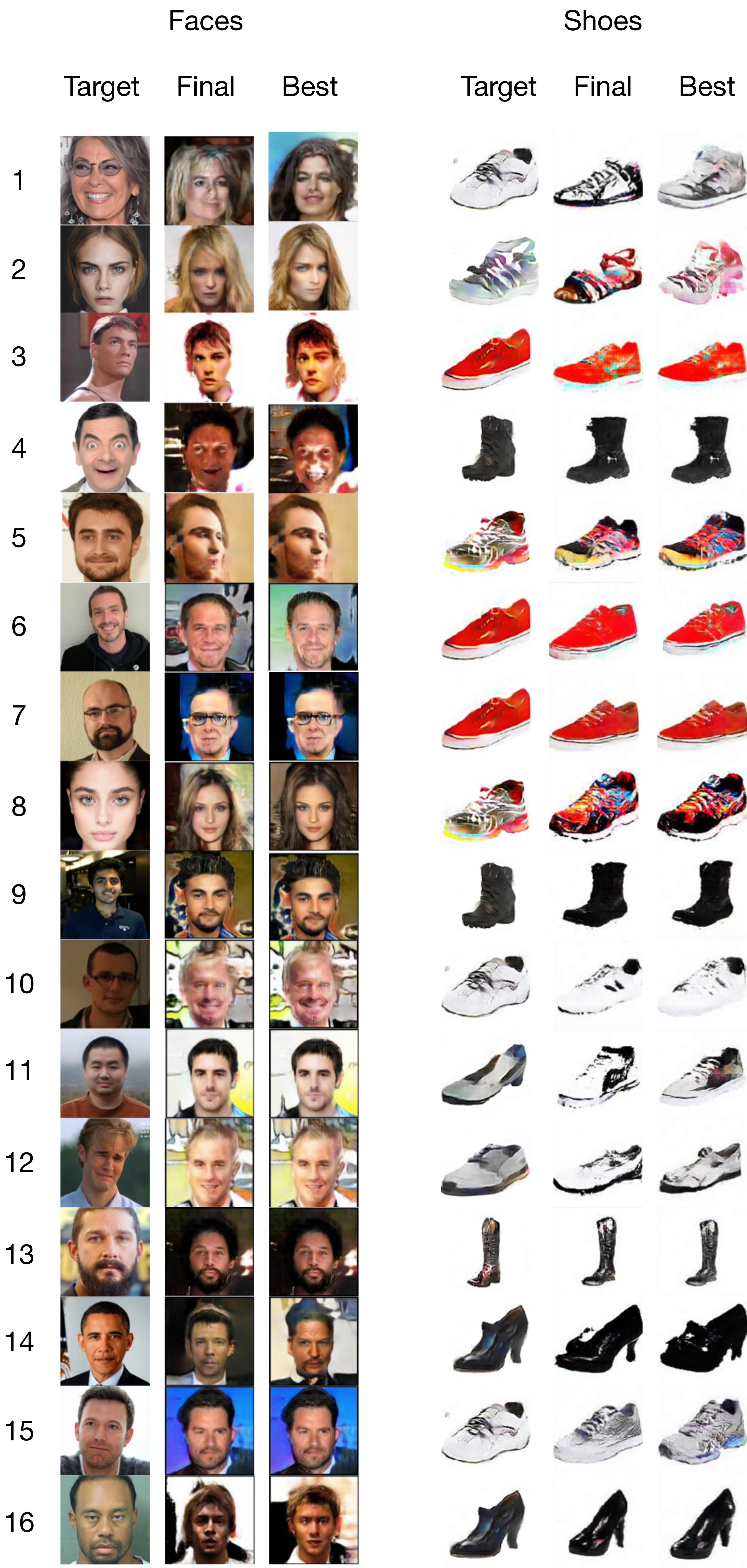} 
  \caption{Evolved Images from the user study. Target images are the intended target for a row. Final images are the last selected image. Best images are the images that the user selected as best from a compilation of all their selections over the entire evolutionary process.} 
  \label{fig:sample}
  \vspace{-1.0em}
\end{center}
\end{figure}

To test the approach in more depth, 16 volunteers tried the system as has been previously described. The outcomes of their efforts can be seen in Figure~\ref{fig:sample}. The target image for each user appears on the left of their respective domain.

For each user, we looked at two different outcomes: the final evolved image, and the user's favorite evolved image. Ideally, these would be the same image but due to the stochastic nature of the optimization process, the best image is sometimes lost. Other times the image does not really improve for the last few generations, making it hard to discern the best generation. With the shoes, in particular, we observed that the image could be recreated by many users within only a few generations. Since we asked the users to go for 10 generations, it resulted in a lot of images that were essentially the same.

The results in Figure~\ref{fig:sample} suggest that there is a wide diversity in how well people were able to recreate the target. The shoes tended to be pretty close, though people are less picky about the distinction between two similar looking shoes. For the face recreation tasks, people usually were only able to focus on a few characteristics.

\subsubsection{Quantitative Analysis}

\begin{figure}[t]
\begin{center}
  \includegraphics[width=0.8\textwidth]{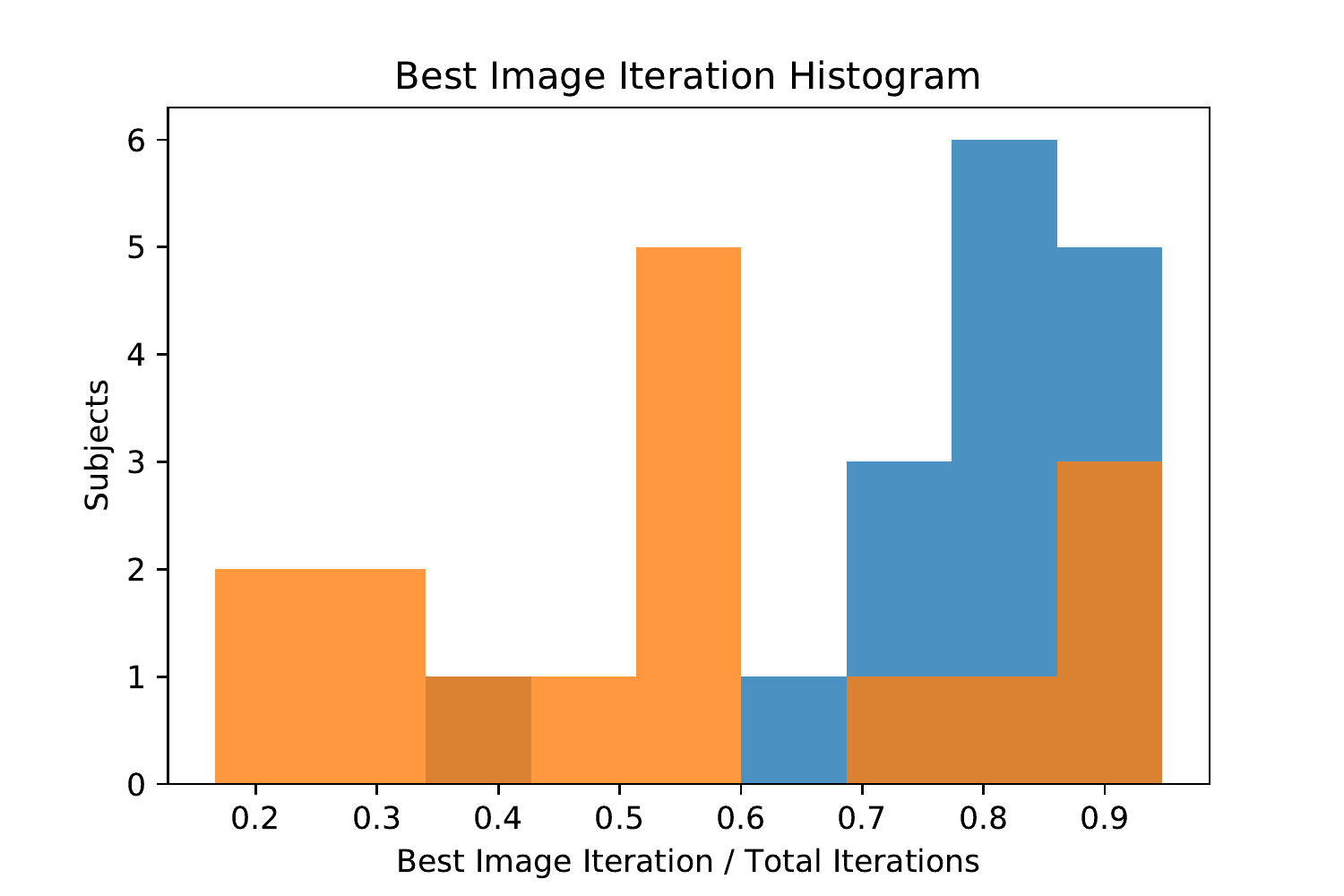} 
  \caption{This histogram depicts at which point during the evolutionary process the best image was found for each of the 16 participants. Data for the shoe domain is shown in orange while the results in the face domain are shown in blue.} 
  \label{fig:hist}
  \vspace{-1.0em}
\end{center}
\end{figure}

As was mentioned earlier, we measured how far into the evolutionary process the best image occurred. Taking this value over the total number of iterations that a user searched provides a ratio of where in the evolutionary process the best image was found. Figure \ref{fig:hist} is a histogram of the distribution of where the best image was found for both shoes and faces. Shoes are represented in orange and faces in blue. It's clear right away that the best faces appeared much closer the final iteration than the best shoes. Looking at a paired-samples t-test, there is a significant difference in the ratios for faces $(\mu=.79, \sigma=.14)$ and shoes $(\mu=.56, \sigma=.16)$ with a $p = 0.001$.

This confirms what was observed from the volunteers, where they said they found the shoe they were looking for in the first few generations, but then continued on to complete the required 10 generations. For this reason, the best shoe image is scattered all over the different generations. As can be seen in figure \ref{fig:sample}, there is often little difference between the best image and final image for shoes. This implies that the search converged early and the shoes mostly looked the same from there on out. The user's "best" pick at the end was mostly a random guess between very similar looking shoes.

The data for faces, on the other hand, tells a different story. The individual ratios are clustered close to 1. This is a good sign that the image of the face was improving throughout the process. The best image was selected at a later time, randomized with all the selected images, and the user still chose one of the later designs. With an average number of generations of 12.7, most of the users stuck to the requested 10 generations so we are not able to measure for how long the faces would continue to improve.

It is important to note this progress is based on the user's reporting of which is the best image. This is by no means an objective measure. This measure of improvement measures that the user is able to better express what they want to express. This is an important metric since this is a communication tool, but the metric shouldn't be confused as an indicator of image quality.

\subsubsection{Experience}

Based on self-reported numbers, users felt that they got much closer to reproducing the shoes than they did to the face. This could be predicted from figure \ref{fig:sample}. On average users reported 2.2 out of 5 for their ability to reproduce faces and 3.8 out of 5 for their ability to reproduce shoes, both with a standard deviation of 1.

What users wrote about their experience was split between amusement and frustration. Some people embrace the chaotic nature of the system while others were frustrated with not having enough control. Part of the frustration could be caused by the nature of the task we assigned them. By providing the image, we gave them a very rigid goal which is different from the intended use case. Of course, our current implementation has limitations in the quality of images that it can produce and that has to be acceptable to the user. 

\subsubsection{Strategies}

By reviewing the user's feedback on the strategies they used, we were able to put together three main strategies. These are the selection strategies people deployed to try and reach their target image. The three primary strategies are as follows;

\begin{enumerate}
  \item Collect Distinct Traits
  \item Select Best Likeness
  \item Hierarchical Trait Selection
\end{enumerate}
Collect Distinct Traits was by far the most popular strategy with over half of the users subscribing to this strategy. This strategy involves breaking down the target into distinct parts and then identifying the parts in the population provided. For a face, this can include hairstyle, gender, face profile, etc. The user of this strategy selects images that are the best representation of these traits. Once traits start to merge into single images, then the individual traits can be ignored. This can be followed until all the desired traits are combined and then the focus moves to finding the best variation of this image.

Select Best Likeness is a simpler strategy. The user looks for the
image or images that are closest to the final result. User 2 in \ref{fig:sample} used this strategy. This strategy is the closest to using a distance-based loss function as the evaluation technique. The main difference is that users reported focusing on the overall look of the image instead of the details which most loss functions are very bad at. A quarter of our subjects reported using this strategy. Half of them implemented it by only selecting one image per generation and relying on the mutation slider.

The final strategy, Hierarchical Trait Selection, is a variation of the first strategy. Instead of maintaining separate threads of different traits, the user focuses on the first desired trait that they see. They then cultivate its diversity until they find an image that has another desired trait. At this point, they have two desired traits and are ready to repeat the process for adding a third. This was the least common strategy with only two people using it. User 16, is one of the users who used this strategy.

We did not expect these distinct strategies to emerge, as such, we have no experiment that compares the effectiveness of the different strategies. We also don't measure how practice affects the outcome. These strategies could also just be the result of trying to recreate faces. They do seem like fundamental methods for designing complex systems through a series of selections.

In the deploying of these strategies, users found specific things they wanted from the interface that is not there. The two most popular requests were for some form of elitism and for the ability to re-roll random samples. Elitism would allow people to keep their best images and then feel comfortable to explore more risky strategies. Re-rolling, or generating a new batch of random samples, would allow the users to quickly search for new features without affecting the features they already found.

These emergent strategies and requests show that there are lots of ways to optimize the interface that can be explored. It also demonstrates that even with one of the simplest interfaces there is a certain depth to designing an image through selection.

\section{Discussion}

This work allows interactive evolutionary computation to be brought to many new domains. Existing techniques are either very general, but very difficult to corral towards a specific problem domain, or they require a hand build model of the domain of the problem domain. By building the domain model using data, there are many new domains that IEC can be applied too.

Obviously, there are still many things that cannot be modeled well with these machine learning techniques, but there is a lot of active research in the field and the applications keep growing. Aside from the ability to learns domains directly from data, DeepIEC allows the user to control the direction of the evolutionary process. Despite the complex mapping from latent variables to output, GAN's are trained to have a smooth output. This means two very similar latent codes will very likely produce similar outputs. Also, the midpoint between two latent codes, not too far apart, is likely to map to images that share qualities of both original images. The way that variation and combinations in the phenotype carry over to the genotype allows the evolutionary process be more transparent. This, in turn, makes it possible to direct the process.

The testing that we did on our implementation of DeepIE focused on the ability for the user to guide the population toward a predefined target. This ability is great for communicating an idea or design that you otherwise might not have the ability to create. This allows someone without the ability to draw to better express themselves through the medium of pictures. But we did not look, in this work, at our tool's ability to be used for creative exploration.

DeepIE shares a key strength that IEC has, which is its ability to help a user explore a creative space. The user can simply select the most interesting suggestion and see where it takes them. In fact, some of the volunteers for our system were tempted to do that at times instead of pursuing their target. By coupling this ability to help a user explore a space with DeepIE's ability to focus on a domain, we get a tool that can help a designer come up with new ideas. This is an area we think that DeepIE could have an impact.

While this is a positive for exploration, it is a downside for some applications of communication. For example, if someone were to try to use this to help victims of a crime create a picture of the perpetrator, they wouldn't want the system's suggestions to be influencing the user. That would be a case where the options that come up on the screen would likely bias the user's memory of the criminal. Even if the generator could produce every face with perfect fidelity, extensive tests would have to be done to make sure the process doesn't alter a person's memory.

It is clear that the interface affects what the user discovers. For example, by not allowing users to save their best images, they are more conservative in exploration. By not giving them enough diversity, in phenotypes, they feel powerless to guide the process in a direction they are interested in. This means that the optimal interface might make a big difference in ensuring that the user always feels in control and engaged in the process.

Going forward there are many directions for this. Different generative models and evolution ideas can be experimented with. New and interesting types of domains can be explored, such as time-based domains. Finally, the way the user interacts, as discussed, can itself go in many directions.

\section{Conclusion}

This paper presented a novel approach to interactive evolution. Users are able to interactively evolve the latent code of a GAN that is pre-trained on a certain class of images (e.g.\ shoes, faces, chairs). Our approach tries to strike a balance between a system like Picbreeder that allows the evolution of arbitrary images (but is very hard to guide towards a specific target), and systems like GANs, in which images are normally sampled randomly without any user input.  
We presented initial results that show that users are able to interactively recreate target images through our DeepIE approach, which has been shown difficult with traditional IEC approaches. In the future, it will be interesting to extend the approach to other domains such as video games that can benefit from high-quality and controllable content generation.
\raggedbottom

\pagebreak
\bibliographystyle{splncs03}
\bibliography{bibliography}
\end{document}